\documentclass[acmlarge]{acmart-ecaf}

\renewcommand{\baselinestretch}{1.15}

\AtBeginDocument{%
  }

\setcopyright{none}
\copyrightyear{2026}
\acmYear{2026}
\acmConference[ECAF'26]{Fifth European Conference on Algorithmic Fairness}{September 02-September 04, 2026}{Ghent, BE}
\usepackage[skip=5pt]{caption} 
\usepackage{microtype}
\begin{document}

\title{Sequential Fairness Auditing with Limited Output Access}


\author{Ioannis Pitsiorlas}
\authornote{Both authors contributed equally to this research.}
\affiliation{%
  \institution{EURECOM}
   \country{France}
}
\email{ioannis.pitsiorlas@eurecom.fr}

\author{Martha V. Sourla}
\authornotemark[1]
\affiliation{%
  \institution{DaSCI, University of Granada}
  \country{Spain}
}
\email{msourla@ugr.es}

\author{Marios Kountouris}
\affiliation{%
  \institution{DaSCI, University of Granada}
  \country{Spain}
}
\email{mariosk@ugr.es}

\renewcommand{\shortauthors}{Pitsiorlas et al.}

\begin{abstract}
External evaluations are becoming increasingly central to the governance of AI systems. In practice, however, independent auditors often have limited access to deployed models and must rely on query-based interactions. Most existing fairness evaluation methods assume static datasets and fixed-sample statistical tests, making them poorly suited to real-world auditing scenarios in which evidence must be collected sequentially under query constraints. In this work, we formulate fairness auditing as a tolerance-aware sequential hypothesis-testing problem under limited model output access. We develop a sequential generalized likelihood-ratio framework that allows auditors to accumulate evidence from a finite audit pool and stop once sufficient support for compliance or violation has been obtained. The framework is instantiated for decision-based Statistical Parity and Equal Opportunity audits, and extended to score- and logit-based proxy audits when richer observables are available. Our results show that both the fairness metric and the level of model access significantly affect audit efficiency, and that the benefits of richer output information are not uniform across auditing settings. In particular, richer outputs can substantially reduce the number of queries required for some fairness metrics and operating regimes, while offering limited gains in near-threshold cases. This work provides a practical statistical framework for sequential fairness auditing under realistic deployment constraints.
\end{abstract}

\keywords{sequential hypothesis testing, algorithmic fairness, fairness auditing, limited output access, proxy audits}

\maketitle

\section{Introduction}
Algorithmic decision systems are increasingly subject to regulatory
oversight. Frameworks such as the EU AI Act \cite{eu2024aiact} and
local bias audit laws \cite{nycll144} require statistical evidence
of compliance for deployed artificial intelligence (AI) systems.
In practice, however, external auditors rarely have access to
model parameters, training data, or internal documentation.
Instead, auditing often occurs through query-based interaction with deployed systems, where auditors observe outputs for selected inputs and must draw conclusions from limited evidence
\cite{hartmann2026auditcanqueryefficientactive,cen24ACM}. In this paper, we distinguish between exact black-box audits of decision-based fairness constraints and access-aware proxy audits that use richer observables, such as scores or logits, to study how additional observability affects sequential evidence accumulation.

Most current fairness evaluation methods rely on static disparity
metrics such as Statistical Parity (SP) or Equal Opportunity (EO)
\cite{Hardt16Equal, Feldman15demo} and assess compliance using
fixed-sample hypothesis tests
\cite{Taskesen21StatTestFairness, Yik22IdentifyingBias}.
These approaches are typically designed for batch evaluation on
pre-collected datasets.
In contrast, real-world audits often involve sequential and query-efficient evidence collection under limited query budgets, where auditors adaptively gather observations and must decide whether sufficient evidence has been obtained to certify compliance or detect violations \cite{yan22c}.

Consider an external auditor evaluating a 
credit approval system for compliance with fairness regulations. The auditor does not have access to the model's parameters or training data and instead interacts with the system through
queries, submitting input profiles, and observing the model's decisions.
Suppose the regulator requires that the disparity in approval rates between demographic groups does not exceed a tolerance threshold $\delta$.
Because each query may correspond to a costly or limited audit interaction, the auditor seeks to determine compliance using as few queries as possible.
This setting naturally leads to a sequential statistical decision problem: as evidence accumulates, the auditor must
decide whether sufficient evidence exists to certify compliance, detect a violation, or continue collecting observations.

Recent work has emphasized auditing under limited access to deployed systems, often framing auditing as a statistical decision problem over observable outputs \cite{Raji20Closing, cen24ACM}. However, existing approaches do not provide concrete sequential procedures with explicit stopping rules for efficient auditing under query constraints.

Sequential hypothesis testing provides principled tools for decision making under incremental data collection, enabling early stopping with controlled error rates \cite{wald1947sequential, Howard21timeuni}. However, to the best of our knowledge, these methods have not been developed into a concrete generalized likelihood ratio (GLR)-based framework for tolerance-aware fairness auditing under limited access.

In this work, we formalize fairness auditing as a tolerance-aware
sequential hypothesis testing problem.
Given a fairness metric $g(f)$ and regulatory tolerance $\delta$,
we consider the composite hypothesis test
\[
H_0: g(f) \le \delta
\qquad
\text{vs.}
\qquad
H_1: g(f) > \delta.
\]
To address this problem, we adopt a GLR framework with sequential stopping boundaries inspired by
Wald-type tests.
Unlike fixed-sample procedures, the proposed protocol allows
evidence to accumulate sequentially and terminates once sufficient
statistical support is obtained or a predefined query budget is
exhausted.
This formulation reflects realistic regulatory settings in which
small disparities may be legally permissible, and compliance is
defined relative to a tolerance threshold rather than zero disparity.

Our empirical study reveals several important phenomena in
sequential fairness auditing.
First, tolerance-aware auditing exhibits strong boundary effects:
models whose fairness gaps lie near the regulatory tolerance
often yield inconclusive outcomes within realistic query budgets.
Second, richer output-access regimes can substantially reduce audit cost for SP and for some operating regimes, but the gains are not uniform across fairness criteria. In particular, the EO results show that richer observables do not automatically yield faster or more conclusive audits near the regulatory threshold.
Third, EO audits are significantly more sample
intensive than SP audits because the EO procedure operates on a conditional $Y=1$ audit pool that is typically smaller, and may therefore provide fewer informative observations and less favorable group counts than the full SP pool.

These findings highlight that fairness audit outcomes depend not only on the deployed model but also on the statistical protocol and level of access available to the auditor.
By operationalizing fairness auditing as a sequential inference problem, our framework provides a foundation for analyzing the reliability and efficiency of real-world AI audits under realistic
deployment constraints.

\subsection{Contributions}
Our contributions are summarized as follows: 
(1) \textbf{Sequential formulation of fairness auditing:} We formulate fairness auditing under limited model output access as a sequential statistical decision problem, where evidence is collected through query-based interaction and compliance is defined relative to a tolerance threshold. 
(2) \textbf{Sequential GLR testing protocol:} We develop a sequential GLR audit with explicit stopping rules and operational decision thresholds for composite fairness constraints. 
(3) \textbf{Access-aware auditing analysis:} We study how different access regimes affect the efficiency of fairness audits, distinguishing between decision-only audits that directly target hard-decision SP/EO estimands and proxy audits based on scores or logits when richer observables are available. 
(4) \textbf{Empirical study:} Through experiments on benchmark datasets, we identify practical phenomena in sequential fairness auditing, including increased difficulty for conditional metrics such as EO and statistical ambiguity near regulatory thresholds.

\section{Related Work}

\subsection{Statistical Fairness Evaluation}

Prior work studies algorithmic fairness through statistical disparity metrics that compare predictive behavior across demographic groups. Some examples are SP, which requires equal positive prediction rates across groups, and EO, which requires equal true positive rates conditioned on the ground truth \cite{Hardt16Equal,Feldman15demo}. These metrics and their variants are widely used to quantify group-level disparities in machine learning systems and have become common benchmarks in fairness evaluation.

Several approaches incorporate statistical testing to assess whether the observed group disparities are significant. For example, hypothesis testing frameworks for fairness metrics have been studied in \cite{Taskesen21StatTestFairness, Yik22IdentifyingBias}. These methods typically construct confidence intervals or perform significance tests for fairness metrics computed on fixed datasets.
Most existing approaches operate in batch settings with fixed datasets and point null hypotheses. In contrast, our work considers sequential, tolerance-based auditing under limited query access.

\subsection{Limited-Access Auditing and Sequential Inference}

A growing body of work studies how AI systems can be audited under limited access conditions. In many real-world deployments, external auditors cannot inspect model parameters, training data, or internal decision logic, and must instead rely on query-based interactions with deployed systems. Recent work has begun to formalize the capabilities and limitations of such limited-access auditing. For example, \cite{cen24ACM} frames auditing as hypothesis testing over observable outputs, while \cite{Casper2024} highlights the limitations of purely black-box evaluation and the potential need for additional access to ensure reliable assessments.

Sequential hypothesis testing provides a natural framework for such settings. Classical methods, beginning with Wald \cite{wald1947sequential}, enable early stopping while controlling error rates under incremental data collection. More recent developments in time-uniform inference and confidence sequences extend these ideas to adaptive settings \cite{Howard21timeuni, Ramdas20game}. Our setting differs from standard sequential testing in two key aspects: the null hypothesis is tolerance-based rather than point-valued, and the observation model depends on the auditor's level of access. The use of GLR testing for such composite, access-dependent fairness auditing problems has not been systematically studied.

\section{Methodology}
We study fairness auditing as a sequential statistical inference problem arising from interaction with a deployed classifier through queries. In the empirical design studied in this paper, the auditor interacts with a fixed held-out audit pool and the sequential procedure reveals one pool element at a time. An auditor repeatedly queries the model on inputs taken from that audit pool and observes model outputs that may vary depending on the level of access provided by the system. The objective is to determine whether the deployed model satisfies a tolerance-based fairness constraint while minimizing the number of queries required to reach a reliable decision.
To this end, we formulate the audit as a sequential hypothesis
testing problem and derive likelihood-based statistics that can
operate under different access regimes.

\subsection{Audit Target}

We study the problem of query-based fairness auditing for deployed binary classifier models. An auditor interacts with a fixed model through input-output queries, and may additionally have access to limited side information, specifically model confidence scores or logits, but not model parameters or training data.

Let $f : \mathcal{X} \rightarrow \{0,1\}$ denote a deployed binary classifier. Each input $x \in \mathcal{X}$ is associated with a sensitive attribute $A \in \{a,b\}$ indicating membership in one of two protected groups.

Let $Q$ denote the reference deployment distribution over $(X,A,Y)$. The ground-truth label $Y \in \{0,1\}$, required for EO, is available from a labeled reference pool. Operationally, the empirical study instantiates this reference distribution using a fixed held-out audit pool. Therefore, reported compliance decisions are evaluated on this pool (or its EO-specific subpool), rather than on an unobserved population beyond it.

The audit target is a fairness function evaluated under $Q$, denoted $g_Q(f)$. A model is declared compliant relative to a regulatory tolerance $\delta > 0$ if $g_Q(f) \le \delta$. The primary compliance target throughout the paper is defined for the deployed hard-decision rule $f$. Thus, the decision-only audit directly targets decision-based SP and EO. When richer outputs are available, we additionally consider auxiliary proxy audits based on group disparities in scores or logits. These richer-output audits preserve the same conditioning structure as the underlying fairness criterion (for example, conditioning on $Y=1$ for EO), but they should be interpreted as operational surrogates rather than exact tests of the hard-decision SP/EO constraints.

\subsection{Fairness Definitions}
\label{section3.2}

We instantiate the framework for the fairness metrics of SP \cite{Feldman15demo} and EO \cite{Hardt16Equal}.

\paragraph{Statistical Parity}
Under distribution $Q$, define the group-conditional positive rates:
\begin{equation}
p_g := \Pr_Q(f(X)=1 \mid A=g), 
\qquad g \in \{a,b\}.
\end{equation}
The SP disparity is defined as 
\begin{equation}
g^{\mathrm{SP}}_Q(f) := |p_a - p_b|.
\end{equation}
We consider the tolerance-based compliance hypothesis test
\begin{equation}
H_0^{\mathrm{SP}} : |p_a - p_b| \le \delta
\qquad \text{vs.} \quad
H_1^{\mathrm{SP}} : |p_a - p_b| > \delta.
\end{equation}

\paragraph{Equal Opportunity}
EO conditions on the true label $Y=1$. Define
\begin{equation}
q_g := \Pr_Q(f(X)=1 \mid A=g, Y=1),
\qquad g \in \{a,b\}.
\end{equation}
The EO disparity is
\begin{equation}
g^{\mathrm{EO}}_Q(f) := |q_a - q_b|.
\end{equation}
The corresponding compliance hypothesis test is
\begin{equation}
H_0^{\mathrm{EO}} : |q_a - q_b| \le \delta
\qquad \text{vs.} \qquad
H_1^{\mathrm{EO}} : |q_a - q_b| > \delta.
\end{equation}

\subsection{Sequential Data and Likelihood Models}

For the finite-pool audit used in the experiments, we distinguish between the SP and EO query sequences. Under SP, the auditor works with the full held-out audit pool and, at each round $t$, queries the next instance in a uniformly random permutation of that pool; equivalently, sampling is uniform without replacement. Under EO, inference is carried out on the filtered conditional pool obtained by restricting the held-out audit pool to instances with $Y=1$. The EO query sequence is generated by a uniformly random permutation of this filtered pool, so every queried instance is EO-relevant by construction. Let $\hat{Y}_t = f(X_t)$ be the observed model prediction.

In this finite-pool setting, the Bernoulli likelihoods below are used as operational working models for sequential evidence accumulation on the revealed observations. They are exact for independent Bernoulli sampling, but here they serve as tractable GLR-style models under without-replacement querying.

\paragraph{SP likelihood model (binary groups)}
For SP, the data available at time $t$ is
\begin{equation}
E_t^{\mathrm{SP}} = \{(A_i,\hat{Y}_i):i=1,...,t\}.
\end{equation}
Under the working likelihood model, conditional on the queried sensitive attributes 
$\{A_i\}_{i=1}^t$, the predicted outcomes 
$\{\hat Y_i\}_{i=1}^t$ are modeled as conditionally independent Bernoulli variables with group-specific success probabilities:
\begin{equation}
\hat Y_i \mid (A_i = g)
\sim
\mathrm{Bernoulli}(p_g),
\qquad g \in \{a,b\}.
\end{equation}
Let $n_g(t) := \sum_{i=1}^t \mathbf{1}\{A_i = g\}$ be the number of samples from group $g$ up to time $t$ and $s_g(t) := \sum_{i=1}^t \mathbf{1}\{A_i = g, \hat Y_i = 1\},$ the corresponding number of positive outcomes, where $g \in \{a,b\}$.
\\
Then the conditional likelihood at time $t$ is
\begin{equation}
L_t^{\mathrm{SP}}(p_a,p_b)
=
p_a^{\,s_a} (1-p_a)^{\,n_a-s_a}
\,
p_b^{\,s_b} (1-p_b)^{\,n_b-s_b},
\end{equation}
where, for brevity, $n_g = n_g(t)$ and $s_g = s_g(t)$ for $g \in \{a,b\}$.

\paragraph{EO likelihood model (binary groups)}
Under EO, inference is conditional on $Y=1$ and is therefore formulated on the conditional reference distribution $Q_+$. The data is analogously defined as 
\begin{equation}
E_t^{\mathrm{EO}} = \{(A_i,\hat{Y}_i):i=1,...,t\}, \qquad (X_i,A_i) \sim Q_+.
\end{equation}
Conditional on the queried sensitive attributes, the EO predicted outcomes satisfy
\begin{equation}
\hat Y_i \mid (A_i = g)
\sim
\mathrm{Bernoulli}(q_g),
\qquad g \in \{a,b\}.
\end{equation}
Because the EO audit samples directly from the $Y=1$ pool, the relevant group-wise counts are defined without an additional $Y_i=1$ indicator:
\begin{equation}
m_g(t) := \sum_{i=1}^t \mathbf{1}\{A_i = g\}, \qquad r_g(t) := \sum_{i=1}^t \mathbf{1}\{A_i = g,\, \hat Y_i = 1\}.
\end{equation}
Thus, the EO conditional likelihood is
\begin{equation}
L_t^{\mathrm{EO}}(q_a,q_b)
=
q_a^{\,r_a} (1-q_a)^{\,m_a-r_a}
\,
q_b^{\,r_b} (1-q_b)^{\,m_b-r_b},
\end{equation}
where, again, $m_g = m_g(t)$ and $r_g = r_g(t)$ for $g \in \{a,b\}$.

In the empirical study, EO can be more challenging than SP because it is conducted on a smaller conditional audit pool restricted to $Y=1$, which may reduce the total number of available queries and alter the group composition of the pool.

\subsection{Audit Statistics under Different Access Regimes}

The information available to an auditor depends on the level of
access provided by the deployed model.
We consider three commonly encountered access regimes.

\paragraph{Decision-only access (predictions only)}
Under decision-only access, often regarded as black-box access, the auditor observes only the binary
prediction $\hat{Y}=f(X)\in\{0,1\}$ returned by the model.
This setting corresponds to the classical fairness auditing scenario in which only final decisions are observable. It is also the regime that directly matches the decision-based SP and EO definitions in Section~\ref{section3.2}.

Under this regime, outcomes for group $g\in\{a,b\}$ are modeled as
Bernoulli variables with parameter $\theta_g$:
\begin{equation}
\hat Y_i \mid A_i=g \sim \mathrm{Bernoulli}(\theta_g).
\end{equation}

The sequential GLR statistic therefore uses the Bernoulli likelihood described in the previous section, where the sufficient statistics are the number of samples $n_g(t)$ and the number of positive outcomes $s_g(t)$ observed for each group under SP, and, under EO, the audit is run directly on the conditional $Y=1$ pool, yielding sufficient statistics $m_g(t)$ and $r_g(t)$ defined on that EO pool.

\paragraph{Score access}
In many deployed systems, the model additionally exposes prediction scores or probabilities $S(x)\in[0,1]$, typically corresponding to the sigmoid output of a classifier.
These scores contain more information than binary decisions and can be used directly by the auditor.

Under score access, often referred to as gray-box access, we consider proxy audits based on group-conditional score means. For SP-style proxy auditing, we define
\begin{equation}
\mu_g^{(S)} := \mathbb{E}[S(X) \mid A=g], \qquad g \in \{a,b\},
\end{equation}
and the corresponding proxy disparity as $d_{\mathrm{SP}}^{(S)} := |\mu_a^{(S)}-\mu_b^{(S)}|$. For EO-style proxy auditing, we retain the same conditioning structure as EO and define
\begin{equation}
\mu_{g,+}^{(S)} := \mathbb{E}[S(X) \mid A=g, Y=1], \qquad g \in \{a,b\},
\end{equation}
with proxy disparity $d_{\mathrm{EO}}^{(S)} := |\mu_{a,+}^{(S)}-\mu_{b,+}^{(S)}|$.

Operationally, we model the observed scores as Gaussian with group-specific means and common variance,
\begin{equation}
S_i \mid A_i=g \sim \mathcal{N}(\mu_g^{(S)},\sigma_S^2),
\end{equation}
and analogously under the EO conditioning event $Y=1$. Because these score-based quantities are not identical to hard-decision SP/EO, we use an operational tolerance in score space rather than identifying it with the hard-decision tolerance $\delta$.

\paragraph{Logit access}
In some settings, the auditor may observe the model's pre-sigmoid
logits $Z(x)$ rather than probabilities. Logits provide an unbounded continuous signal that often carries
more information about the classifier's internal confidence (scoring behavior).

Under logit access, sometimes referred to as white-box access, although it does not require access to model parameters, we analogously define proxy audits based on group-conditional logit means. For SP-style proxy auditing, let $\mu_g^{(Z)} := \mathbb{E}[Z(X) \mid A=g]$, and for EO-style proxy auditing let $\mu_{g,+}^{(Z)} := \mathbb{E}[Z(X) \mid A=g, Y=1]$. The corresponding proxy disparities are $d_{\mathrm{SP}}^{(Z)} := |\mu_a^{(Z)}-\mu_b^{(Z)}|$ and $d_{\mathrm{EO}}^{(Z)} := |\mu_{a,+}^{(Z)}-\mu_{b,+}^{(Z)}|$.

We model the observed logits using a Gaussian likelihood with group-specific means and common variance,
\begin{equation}
Z_i \mid A_i=g \sim \mathcal{N}(\mu_g^{(Z)},\sigma_Z^2),
\end{equation}
again conditioning on $Y=1$ for EO-style proxy audits. Importantly, we do not identify a probability-space tolerance with a unique logit-space tolerance, since the logistic transform is nonlinear. Instead, the logit-access audit uses an operational threshold in logit space.

Across regimes, the sequential stopping rule retains the same form; what changes is the observation model and, for richer observables, the disparity being audited. For the score-based and logit-based proxy audits, the Gaussian likelihoods are likewise used as working models on the revealed finite-pool observations rather than as exact without-replacement sampling laws.

\subsection{Sequential Generalized Likelihood Ratio Test}
To operationalize audits as statistical hypothesis tests, we adopt a GLR framework. We formulate the test using GLR because both $H_0$ and $H_1$ define sets of admissible parameter values rather than single distributions.
For the decision-only audits targeting the hard-decision estimand, let $\theta = (\theta_a,\theta_b)$ denote the vector of group-conditional Bernoulli parameters, with
\[
\theta =
\begin{cases}
(p_a, p_b), & \text{for Statistical Parity}, \\
(q_a, q_b), & \text{for Equal Opportunity}.
\end{cases}
\]
Define the disparity functional
\begin{equation}
d(\theta) := |\theta_a - \theta_b|.
\end{equation}
Given tolerance level $\delta > 0$, define the compliance and violation parameter sets
\begin{equation}
\Theta_0(\delta)
=
\{\theta \in [0,1]^2 : d(\theta) \le \delta\},
\end{equation}
\begin{equation}
\Theta_1(\delta)
=
\{\theta \in [0,1]^2 : d(\theta) > \delta\}.
\end{equation}

At time $t$, let $L_t(\theta)$ denote the appropriate likelihood 
(SP or EO as defined above). 
The GLR statistic is
\begin{equation}
\Lambda_t
=
\log
\frac{\sup_{\theta \in \Theta_1(\delta)} L_t(\theta)}
     {\sup_{\theta \in \Theta_0(\delta)} L_t(\theta)}.
\end{equation}
For numerical convenience, it is useful to express the statistic in terms of the log-likelihood $\ell_t(\theta) := \log L_t(\theta).$
Then the GLR statistic can be equivalently written as
\begin{equation}
\Lambda_t
=
\sup_{\theta \in \Theta_1(\delta)} \ell_t(\theta)
-
\sup_{\theta \in \Theta_0(\delta)} \ell_t(\theta).
\end{equation}

For score- and logit-based proxy audits, the same GLR construction is used with $\theta$ replaced by the relevant pair of group-conditional means and with $L_t$ taken to be the Gaussian likelihood under a common variance model. In that case, the null and alternative hypotheses are defined by the corresponding score-space or logit-space proxy disparity, rather than by the hard-decision SP/EO gap. In the Gaussian proxy tests, the common variance is treated as a nuisance parameter and profiled at each step using the pooled residual variance under the corresponding constrained or unconstrained fit. This profiling choice matches the implementation used to generate the reported results.

After $t$ audit queries, the statistic $\Lambda_t$ is updated sequentially and compared with upper and lower stopping boundaries $(u,\ell)$. Because the present setting involves a composite GLR test, the classical simple SPRT thresholds are not claimed here as exact finite-sample guarantees. Instead, the boundaries are treated as operational decision thresholds for the sequential GLR audit. For nominal error levels $(\alpha,\beta)$, we use the SPRT-inspired values $u=\log((1-\beta)/\alpha)$ and $\ell=\log(\beta/(1-\alpha))$. With $\alpha=0.05$ and $\beta=0.2$, this gives $u\approx2.77$ and $\ell\approx-1.56$.

In the experiments below, we use a nominal maximum query budget of $B=4000$ instances. The boundaries $(u,\ell)$ are the fixed operational thresholds above and are held constant across all datasets, models, metrics, and access regimes. Because the present paper focuses on comparative audit behavior rather than exact finite-sample calibration, we report empirical decision frequencies and stopping times instead of claiming exact Type~I/Type~II guarantees. Under the finite-pool implementation, the effective stopping cap is \[T_{\max}^{(m)}=\min\{B,N_{\mathrm{pool}}^{(m)}\},\] where $N_{\mathrm{pool}}^{(m)}$ denotes the size of the audit pool for metric $m$. For SP, $N_{\mathrm{pool}}^{(\mathrm{SP})}$ is the size of the full held-out audit pool, whereas for EO, $N_{\mathrm{pool}}^{(\mathrm{EO})}$ is the size of the filtered $Y=1$ pool and can be smaller than $B$.

The audit rejects $H_0$ if $\Lambda_t \ge u$, accepts $H_0$ if $\Lambda_t \le \ell$, and continues sampling otherwise. If neither boundary is crossed within the maximum audit budget $B$ or before the corresponding finite audit pool is exhausted, the outcome is declared inconclusive.

The constrained maximizations required by the GLR reduce to one-dimensional boundary optimizations over the tolerance constraint. The full derivation of the unconstrained and constrained maximum likelihood estimates is provided in Appendix~\ref{app:glr_details}.

\section{Experimental Setup}
Our experiments evaluate whether the proposed sequential auditing procedure can reliably distinguish between compliant and non-compliant models under different auditor observability regimes. In particular, we study how audit cost changes as the auditor gains progressively richer access to model outputs, ranging from decision-only access to score-level and logit-level access.

\subsection{Datasets Used}

We evaluate our auditing framework on two standard tabular datasets commonly used in fairness studies: the American Community Survey (ACS) and the Adult dataset.

\textbf{ACS (Folktables).}  
We use the ACS Income task from the Folktables benchmark suite~\cite{ding2021retiring}, which predicts whether an individual's income exceeds \$50,000 based on demographic and employment attributes. We focus on data from California (2018, 1-Year survey). The protected attribute is \textbf{sex}, with two groups: male and female.

\textbf{Adult (UCI).}  
We also evaluate on the Adult dataset from the UCI repository, which has the same prediction objective (income above \$50,000) and is widely used as a benchmark in fairness research. We again use \textbf{sex} as the protected attribute, defining male and female groups.

For both datasets, we apply standard preprocessing, including one-hot encoding of categorical features and standardization of numerical variables. Each dataset is split into training and test partitions, and the held-out test partition serves as the audit pool.

\subsection{Models}

To study how audit behavior depends on model capacity and training stability,
we construct four audited models based on multilayer perceptrons.

\paragraph{Base Architectures}
All base models are implemented as feedforward neural networks
with ReLU activations and a sigmoid output layer.
The \textbf{Robust-Base} model consists of two hidden layers
with 64 and 32 units respectively,
trained for 30 epochs using the Adam optimizer \cite{kingma2015adam} with learning rate $10^{-3}$ and dropout rate 0.1.
The \textbf{Unstable-Base} model consists of a single hidden layer
with 16 units, trained for 1 epoch
with learning rate $10^{-2}$ and no regularization.
This configuration induces higher variance and underfitting,
leading to less stable decision boundaries.

\paragraph{Fairness-Constrained Models}
To obtain fairness-aware variants, we apply the Exponentiated Gradient reduction framework with a Demographic Parity constraint \cite{agarwal2018reductions,fairlearn2020}. The \textbf{Robust-DP} model is trained with tolerance parameter $\epsilon = 0.02$ and up to 10 reduction iterations, while the \textbf{Unstable-DP} model uses $\epsilon = 0.05$ and up to 2 reduction iterations. All models are treated as fixed deployed systems during auditing. The auditor does not have access to model parameters or training data.

\subsection{Auditor Access Regimes}

We consider three observability regimes that determine what information the auditor can access from the deployed model.

\begin{itemize}
\item \textbf{Decision-only access}: the auditor observes only binary predictions $f(X) \in \{0,1\}$ and therefore performs the hard-decision SP/EO audit described in Section~\ref{section3.2}.
\item \textbf{Score access}: the auditor observes prediction scores $p(X)$ and performs score-based proxy audits with the same SP-style or EO-style conditioning structure.
\item \textbf{Logit access}: the auditor observes the model logits $z(X)$ prior to the final sigmoid activation and performs the analogous logit-based proxy audits.
\end{itemize}

These regimes allow us to study how additional model information affects the efficiency of sequential auditing.

Across all experiments, we use a nominal maximum budget of $B=4000$ queried instances using the fixed operational GLR thresholds defined in Section 3.5.

\subsection{Fairness Criteria}

We instantiate the auditing framework using two group fairness criteria.

\paragraph{Statistical Parity}
SP requires that the probability of a positive outcome be approximately equal across protected groups. For a sensitive attribute $A$ with values in $\{a,b\}$:
\begin{equation}
\left| \Pr(f(X)=1 \mid A=a) - \Pr(f(X)=1 \mid A=b) \right| \leq \delta .
\end{equation}

\paragraph{Equal Opportunity}
EO requires equality of true positive rates across groups:
\begin{equation}
\left| \Pr(f(X)=1 \mid Y=1, A=a) - \Pr(f(X)=1 \mid Y=1, A=b) \right| \leq \delta .
\end{equation}

For EO auditing, the sequential test is applied only to instances that satisfy $Y=1$. Operationally, we implement this by constructing an EO-specific audit pool consisting only of positively labeled instances and then sampling directly from that pool.

Throughout the experiments, the oracle compliance labels are always computed from the hard-decision SP/EO gaps in Table~\ref{tab:oracle_gaps}. For score and logit access, the sequential procedures are therefore interpreted as proxy audits that preserve the same SP-style or EO-style conditioning structure but operate on scores or logits rather than on hard decisions.
 
\subsection{Audit Sampling Procedure}

All experiments use uniform random sampling without replacement from a finite audit pool. For each run, the relevant audit pool is randomly shuffled, and instances are queried in that order until a stopping boundary is crossed or the available pool is exhausted. This design allows us to isolate the effect of access regime and audit statistic on stopping behavior without introducing adaptive sampling effects. For SP, the audit pool is the full held-out test partition. For EO, we first filter the held-out test partition to the subset with $Y=1$ and then perform sampling directly from this conditional EO pool. Accordingly, the EO procedure operates on a smaller finite population and every EO query is informative by construction. For reproducibility, the number of queries in each audit is bounded by $\min(B,N_{\mathrm{pool}})$, where $B$ is the nominal query budget and $N_{\mathrm{pool}}$ is the size of the available audit pool. For SP, $N_{\mathrm{pool}}$ corresponds to the full held-out test set, while for EO it corresponds to the subset of positively labeled instances. Group-wise counts in these pools determine how quickly the GLR statistic accumulates evidence.

\subsection{Population-Level Fairness Evaluation}

To interpret sequential audit outcomes, we compute population-level fairness gaps using the full test dataset. Specifically, we compute the hard-decision SP and EO gaps for each model across the entire test pool.

These values provide oracle reference labels indicating whether each model is compliant or violating the tolerance $\delta$. Importantly, this information is not available to the auditor during the auditing process and is used only for post hoc evaluation of audit correctness.

Each experimental configuration is evaluated over 20 independent runs with different random seeds. Reported statistics correspond to the mean and standard deviation across these runs.

\section{Numerical Results}

Before evaluating sequential audit behavior, we compute the fairness gaps of all models on the full test datasets. These values serve as an empirical reference indicating whether each model is compliant or non-compliant with respect to the tolerance $\delta$. The auditor does not have access to these quantities. They are used only for post hoc evaluation of audit correctness. Table~\ref{tab:oracle_gaps} reports the population-level fairness gaps for both datasets for $\delta=0.05$.

\begin{table}[t]
\centering
\caption{Population-level fairness gaps on the test datasets.}
\label{tab:oracle_gaps}
\begin{tabular}{l l c c}
\toprule
Dataset & Model & SP Gap & EO Gap \\
\midrule
ACS & Robust-DP     & 0.0287 & 0.0613 \\
ACS & Unstable-DP   & 0.0817 & 0.0011 \\
ACS & Robust-Base   & 0.0923 & 0.0202 \\
ACS & Unstable-Base & 0.0894 & 0.0085 \\

Adult & Robust-DP     & 0.0043 & 0.3233 \\
Adult & Unstable-DP   & 0.1424 & 0.0211 \\
Adult & Robust-Base   & 0.1716 & 0.0753 \\
Adult & Unstable-Base & 0.1706 & 0.0819 \\
\bottomrule
\end{tabular}
\end{table}

With tolerance $\delta = 0.05$, compliance depends on both the fairness metric and the dataset. On ACS, Robust-DP is compliant under SP but violates EO, whereas Unstable-DP violates SP but remains compliant under EO. On Adult, Robust-DP is also compliant under SP but strongly violates EO, while Unstable-DP violates SP and remains compliant under EO. These oracle gaps provide the reference labels used to interpret sequential audit outcomes. To keep the access-regime comparison focused, Tables~\ref{tab:sp_results}--\ref{tab:error_sanity} report the base models, while Table~\ref{tab:oracle_gaps} retains the fairness-constrained models for context.

\subsection{Sequential Auditing Under Different Access Regimes}

We evaluate how the level of access to model outputs affects the efficiency of sequential auditing. We consider three access regimes:
\begin{itemize}
\item \textbf{Decision-only:} the auditor observes only binary predictions.
\item \textbf{Score access:} the auditor observes prediction scores.
\item \textbf{Logit access:} the auditor observes logits before the sigmoid layer.
\end{itemize}

All experiments use shuffled finite audit pools without replacement. Each configuration is evaluated over \textbf{20 independent audit runs} with different random seeds. Reported results correspond to the mean and standard deviation of the number of queried instances required for the sequential test to terminate or for the relevant audit pool to be exhausted.

\subsection{Statistical Parity}

\begin{table*}[t]
\centering
\setlength{\tabcolsep}{4pt} 
\caption{Sequential auditing under SP. The oracle column reports hard-decision SP gaps. Decision-only access corresponds to the hard-decision audit, while score and logit access denote proxy audits. Entries report the decision and mean queries to termination ($\pm$ std).}
\label{tab:sp_results}
\begin{tabular}{l l c c c c}
\toprule
\textbf{Dataset} & \textbf{Model} & \textbf{Oracle SP Gap} & \textbf{Decision-only (labels)} & \textbf{Score access (scores)} & \textbf{Logit access (logits)} \\
\midrule
ACS & Robust-Base   & 0.0923 & Reject, $2349 \pm 1345$ & Reject, $2157 \pm 1443$ & Reject, $708 \pm 543$ \\
ACS & Unstable-Base & 0.0894 & Reject, $2412 \pm 1587$ & Reject, $2406 \pm 1477$ & Reject, $691 \pm 595$ \\

Adult & Robust-Base   & 0.1716 & Reject, $191 \pm 215$ & Reject, $94 \pm 75$ & Reject, $77 \pm 54$ \\
Adult & Unstable-Base & 0.1706 & Reject, $215 \pm 178$ & Reject, $114 \pm 83$ & Reject, $75 \pm 50$ \\
\bottomrule
\end{tabular}
\end{table*}

As shown in Table~\ref{tab:sp_results}, the access-regime decisions agree empirically with the non-compliant hard-decision oracle labels for the base models. For score and logit access, these results should be interpreted as proxy audits relative to the hard-decision SP oracle. However, the number of required queries varies substantially with the level of access. Decision-only auditing consistently requires the most samples, score access reduces this cost by exploiting confidence information, and logit access provides the strongest gains.

The access effect is particularly pronounced on ACS, where the SP gaps of the base models are relatively close to the tolerance threshold. For example, on Robust-Base, the mean sample requirement drops from $2349$ under decision-only access to $708$ under logit access. A similar pattern appears for Unstable-Base, where the mean decreases from $2412$ to $691$. Thus, on ACS, logit access reduces audit cost by roughly a factor of three relative to prediction-only access.

The same qualitative trend holds on Adult. There, the SP violations are substantially larger, so all access regimes terminate much more quickly. Even in this easier setting, richer access remains beneficial: for Robust-Base, the mean sample requirement decreases from $191$ under decision-only access to $94$ under score access and $77$ under logit access; for Unstable-Base, the corresponding values are $215$, $114$, and $75$. These results indicate that richer model outputs consistently improve SP audit efficiency, although the absolute gain depends on how far the true fairness gap lies from the regulatory boundary.

\begin{figure}[t]
\centering
\includegraphics[width=0.85\linewidth]{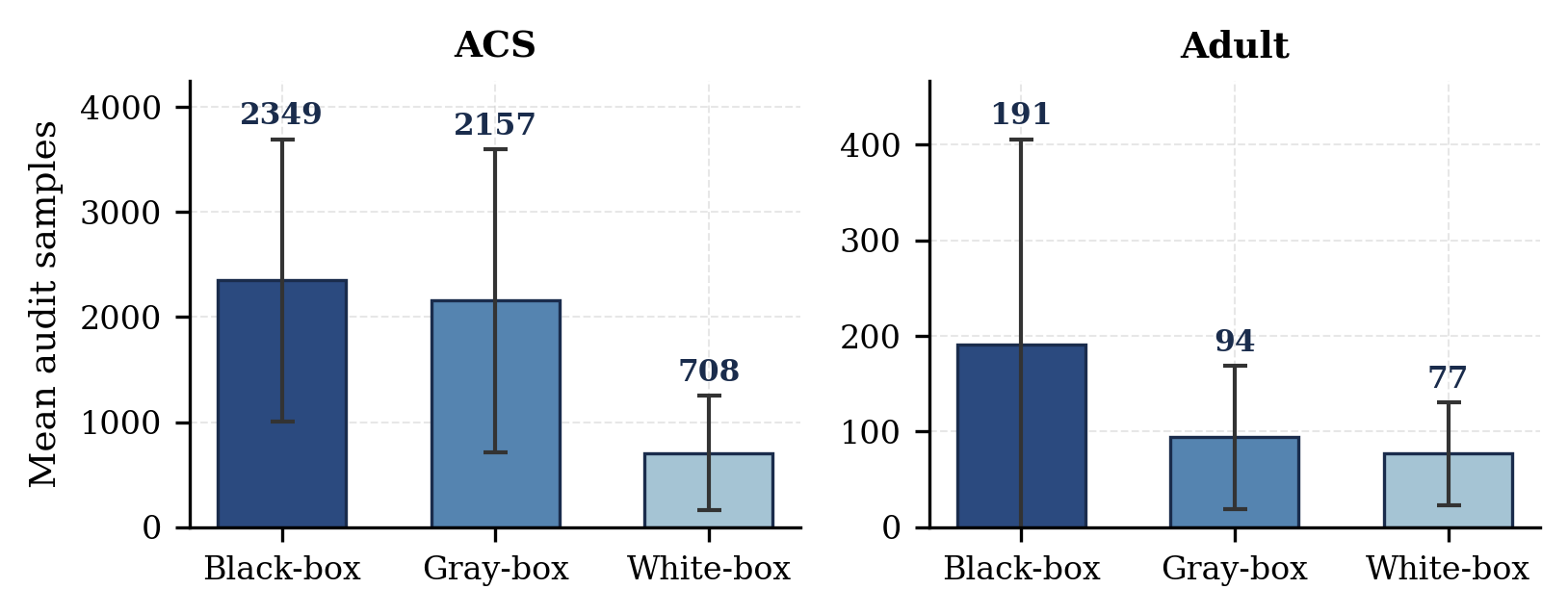}
\caption{Sequential audit cost under SP across access regimes on the ACS and Adult datasets. Panels are shown separately. Error bars denote one standard deviation over 20 runs. Richer access reduces query cost in both datasets.}
\label{fig:sp_access}
\end{figure}

Figure~\ref{fig:sp_access} visualizes these trends for both datasets.

\subsection{Equal Opportunity}

\begin{table*}[t]
\centering
\setlength{\tabcolsep}{4pt} 
\caption{Sequential auditing under EO. Oracle values are hard-decision EO gaps. Decision-only access denotes the hard-decision audit; score and logit access denote proxy audits conditioned on $Y=1$. Entries report decision and mean queries ($\pm$ std). ``Inc.'' (inconclusive) indicates no boundary crossing within the budget or EO pool}.
\label{tab:eo_results}
\begin{tabular}{l l c c c c}
\toprule
\textbf{Dataset} & \textbf{Model} & \textbf{Oracle EO Gap} & \textbf{Decision-only (labels)} & \textbf{Score access (scores)} & \textbf{Logit access (logits)} \\
\midrule
ACS & Robust-Base   & 0.0202 & Accept, $2711 \pm 1123$ & Accept, $1060 \pm 1098$ & Inc., $3329 \pm 1214$ \\
ACS & Unstable-Base & 0.0085 & Accept, $1257 \pm 702$ & Accept, $550 \pm 375$ & Inc., $2829 \pm 1487$ \\

Adult & Robust-Base   & 0.0753 & Inc., $2751 \pm 1224$ & Inc., $2757 \pm 1220$ & Inc., $2512 \pm 1334$ \\
Adult & Unstable-Base & 0.0819 & Inc., $2492 \pm 1340$ & Inc., $3045 \pm 952$ & Inc., $2717 \pm 1171$ \\
\bottomrule
\end{tabular}
\end{table*}

Compared to SP, EO auditing is substantially more challenging, as shown in Table~\ref{tab:eo_results}. In our implementation, this is not because non-positive instances are queried and then discarded, but because EO is conducted directly on the smaller conditional audit pool restricted to $Y=1$. This smaller pool can limit the total number of available queries and can also lead to less favorable group counts for evidence accumulation.

This effect is already visible on ACS. Both base models are EO-compliant, and the decision-only and score-access auditors accept them in most runs, but require many more queries than under SP. For Robust-Base, the mean sample cost is $2711$ under decision-only access and $1060$ under score access, whereas the logit-access audit remains inconclusive on average within the available budget. A similar pattern holds for Unstable-Base, with decision-only and score access reaching acceptance at $1257$ and $550$ samples, respectively, while logit access again frequently fails to terminate conclusively.

EO auditing is even more difficult on Adult. Both base models violate EO, but only moderately relative to the threshold, with oracle gaps of $0.0753$ and $0.0819$. In this near-boundary regime, all three access levels frequently terminate inconclusively within the audit budget. Mean sample counts remain high across all regimes, ranging from approximately $2492$ to $3045$, with no consistent efficiency gain from richer access. This indicates that, for EO, access to scores or logits does not automatically translate into faster audits when the fairness gap is not sufficiently separated from the tolerance level.

Taken together, these results show that access-dependent gains are not metric-independent. Richer access consistently improves performance for the SP task in our experiments, while EO remains intrinsically harder because it conditions on positive labels. In our implementation, this additional difficulty arises from the smaller conditional EO audit pool and its group composition, rather than from post hoc discarding of sampled non-positive instances. The EO score- and logit-access results should therefore be interpreted cautiously as proxy-audit behavior rather than as evidence that richer access uniformly improves fairness auditing.

\subsection{Sequential Versus Fixed-Sample Auditing}

Table~\ref{tab:seq_vs_fixed} compares the sequential GLR auditor with a fixed-sample baseline that uses the full query budget. For comparability, this table reports the decision-only hard-decision audit only. Accuracy treats inconclusive sequential outcomes as incorrect relative to the oracle label. The fixed-sample baseline evaluates the same hard-decision disparity after the full available budget, using $B$ queries for SP and $\min(B,N_{Y=1})$ queries for EO. Across both datasets, sequential auditing substantially reduces the number of required queries compared to the fixed-sample approach. This reduction is particularly pronounced on the Adult dataset under SP, where decisions are obtained with fewer than 250 samples on average, compared to the full budget of 4000.

However, this efficiency comes with a reduction in decision accuracy in more challenging settings. In particular, under EO on the Adult dataset, the sequential auditor achieves significantly lower accuracy than the fixed-sample baseline, reflecting the difficulty of near-threshold cases and limited effective sample sizes.

These results highlight a fundamental efficiency-accuracy trade-off: sequential auditing can dramatically reduce query cost, but may require larger budgets or repeated runs to match the reliability of fixed-sample testing.

\begin{table*}[t]
\centering
\caption{Sequential vs. fixed-sample auditing for the decision-only hard-decision auditor. Accuracy is measured relative to the oracle decision, with inconclusive outcomes counted as incorrect. For EO, the fixed-sample count is $\min(B,N_{Y=1})$, which equals 3362 for Adult.}
\label{tab:seq_vs_fixed}
\begin{tabular}{l l c c c c}
\toprule
\textbf{Dataset} & \textbf{Model} & \textbf{Metric} & \textbf{Method} & \textbf{Accuracy} & \textbf{Samples} \\
\midrule
ACS & Robust-Base   & SP  & Sequential & 0.75 & $2349 \pm 1345$ \\
ACS & Robust-Base   & SP  & Fixed      & 1.00 & $4000$ \\
ACS & Unstable-Base & SP  & Sequential & 0.70 & $2412 \pm 1587$ \\
ACS & Unstable-Base & SP  & Fixed      & 0.95 & $4000$ \\

ACS & Robust-Base   & EO  & Sequential & 0.70 & $2711 \pm 1123$ \\
ACS & Robust-Base   & EO  & Fixed      & 0.95 & $4000$ \\
ACS & Unstable-Base & EO  & Sequential & 0.90 & $1257 \pm 702$ \\
ACS & Unstable-Base & EO  & Fixed      & 1.00 & $4000$ \\

Adult & Robust-Base   & SP  & Sequential & 1.00 & $191 \pm 215$ \\
Adult & Robust-Base   & SP  & Fixed      & 1.00 & $4000$ \\
Adult & Unstable-Base & SP  & Sequential & 1.00 & $215 \pm 178$ \\
Adult & Unstable-Base & SP  & Fixed      & 1.00 & $4000$ \\

Adult & Robust-Base   & EO  & Sequential & 0.20 & $2751 \pm 1224$ \\
Adult & Robust-Base   & EO  & Fixed      & 1.00 & $3362$ \\
Adult & Unstable-Base & EO  & Sequential & 0.30 & $2492 \pm 1340$ \\
Adult & Unstable-Base & EO  & Fixed      & 1.00 & $3362$ \\
\bottomrule
\end{tabular}
\end{table*}

\subsection{Empirical Sanity Check of Decision Behavior}

\begin{table*}[t]
\centering
\caption{Empirical decision behavior of the sequential auditor for a decision-only hard-decision setting. Entries report frequencies across 20 runs.}
\label{tab:error_sanity}
\begin{tabular}{l l c c c c c}
\toprule
\textbf{Dataset} & \textbf{Model} & \textbf{Metric} & \textbf{Oracle} & \textbf{Reject} & \textbf{Accept} & \textbf{Inconclusive} \\
\midrule
ACS & Robust-Base   & SP & Reject & 0.75 & 0.00 & 0.25 \\
ACS & Unstable-Base & SP & Reject & 0.70 & 0.00 & 0.30 \\
ACS & Robust-Base   & EO & Accept & 0.00 & 0.70 & 0.30 \\
ACS & Unstable-Base & EO & Accept & 0.10 & 0.90 & 0.00 \\

Adult & Robust-Base   & SP & Reject & 1.00 & 0.00 & 0.00 \\
Adult & Unstable-Base & SP & Reject & 1.00 & 0.00 & 0.00 \\
Adult & Robust-Base   & EO & Reject & 0.20 & 0.00 & 0.80 \\
Adult & Unstable-Base & EO & Reject & 0.30 & 0.00 & 0.70 \\
\bottomrule
\end{tabular}
\end{table*}

As shown in Table~\ref{tab:error_sanity}, the decision-only sequential auditor rarely makes incorrect decisions, with most uncertainty expressed as inconclusive outcomes. This is especially visible under EO on the Adult dataset, where the auditor often fails to terminate conclusively because the effective sample size is smaller and the fairness gaps are close to the threshold. Although the ACS Unstable-Base EO case yields occasional erroneous rejections, the overall pattern suggests conservative behavior: when evidence is weak, the auditor is more likely to remain inconclusive than to systematically misclassify the model.

\section{Discussion}

Our results highlight several implications for the design and interpretation of real-world fairness audits. First, audit efficiency depends strongly on the level of access available to the auditor. While regulatory settings often assume decision-only or otherwise limited access, our experiments show that richer outputs, such as scores or logits, can substantially reduce the number of queries required to reach a decision for some tasks, particularly under SP. However, these gains are not uniform across settings and should be interpreted as proxy-audit behavior rather than exact tests of decision-based fairness. Second, tolerance-aware auditing introduces inherent statistical ambiguity near regulatory thresholds. When the true disparity lies close to the tolerance level, sequential audits often require many samples and may terminate inconclusively within realistic budgets. This highlights a fundamental tension between regulatory thresholds and statistical detectability. Third, the choice of a fairness metric plays a central role in audit complexity. EO is consistently more challenging than SP in our experiments. In our implementation, this is because EO operates on a smaller conditional audit pool restricted to $Y=1$, which reduces the number of available queries and may yield less favorable group counts for evidence accumulation. More broadly, these findings demonstrate that fairness audit outcomes depend not only on the deployed model but also on the statistical testing protocol and the auditor's level of access.

This work also has several limitations. We focus on binary-group fairness in binary classification, leaving multi-group and multiclass settings to future work. The score- and logit-access settings rely on proxy disparities defined on scores and logits, and therefore do not provide exact compliance guarantees for decision-based SP/EO. The sequential GLR thresholds are treated as operational rather than exact finite-sample calibrations. Finally, the empirical study is conducted on finite audit pools sampled without replacement, while the Bernoulli and Gaussian likelihoods are used as tractable working models within the sequential procedure.
Future work could explore extensions to multi-group fairness settings, adaptive query strategies for improving audit efficiency, and alternative sequential inference tools such as confidence sequences.

\section*{Acknowledgments}

This work was supported by the European Research Council (ERC) under the European Union’s Horizon 2020 research and innovation programme (Grant agreement No. 101003431) and by Horizon Europe/JU SNS project, ROBUST-6G (Grant agreement No. 101139068).

\bibliographystyle{ACM-Reference-Format}
\bibliography{sample-base}

\appendix
\section{GLR Optimization}
\label{app:glr_details}

\subsection*{Unconstrained and Constrained Maximum Likelihood Estimates (MLE)}

\paragraph{Unconstrained MLE}

Let $\hat\theta_t = (\hat\theta_{a,t},\hat\theta_{b,t})$ denote the empirical rate vector at time $t$.
The unconstrained MLE is 
\begin{equation}
\hat\theta_t
=
\left(
\frac{s_a}{n_a},
\frac{s_b}{n_b}
\right)
\text{ (SP)},
\qquad
\hat\theta_t
=
\left(
\frac{r_a}{m_a},
\frac{r_b}{m_b}
\right)
\text{ (EO)},
\end{equation}
with the convention that the test proceeds only after each group has at least one relevant observation so that the denominators are nonzero. This maximizes $L_t(\theta)$ over the parameter space $[0,1]^2$.

\paragraph{Constrained MLE under $H_0$}

If the empirical disparity satisfies $|\hat{\theta}_a-\hat{\theta}_b|\le\delta$, then the unconstrained MLE already lies in $\Theta_0(\delta)$ and therefore also maximizes the likelihood under $H_0$.

Otherwise, the maximizer must lie on the boundary $|\theta_a-\theta_b|=\delta$, which can be parameterized as
\begin{equation}
\theta_b=\theta_a+s\delta, \qquad s\in\{-1,+1\}.
\end{equation}
Substituting this relation into the log-likelihood yields the one-dimensional function
\begin{equation}
f_s(\theta_a)
=
\ell_t(\theta_a,\theta_a+s\delta).
\end{equation}
The feasible values of $\theta_a$ are those satisfying
\begin{equation}
0 \le \theta_a \le 1,
\qquad
0 \le \theta_a+s\delta \le 1.
\end{equation}

For the Bernoulli likelihood (SP case), the derivative is

\begin{equation}
\frac{d}{d\theta_a} f_s(\theta_a)
=
\frac{s_a}{\theta_a}
-
\frac{n_a-s_a}{1-\theta_a}
+
\frac{s_b}{\theta_a+s\delta}
-
\frac{n_b-s_b}{1-\theta_a-s\delta}.
\end{equation}

An analogous expression holds for EO by replacing $(n_g,s_g)$ with $(m_g,r_g)$.

Since the second derivative is always negative, the Bernoulli log-likelihood is strictly concave, the function $f_s(\theta_a)$ is strictly concave over its feasible domain, and therefore admits a unique maximizer for each $s \in \{-1,+1\}$. The constrained MLE $\sup_{\theta \in \Theta_0(\delta)} \ell_t(\theta)$ is obtained by computing the maximizer for both boundary cases and selecting the solution that yields the larger likelihood value.

When the unconstrained MLE lies in $\Theta_1(\delta)$, the GLR statistic can be written as

\begin{equation}
\Lambda_t
=
\ell_t(\hat{\theta}_t)
-
\sup_{\theta \in \Theta_0(\delta)} \ell_t(\theta).
\end{equation}

By symmetry, when the unconstrained MLE lies in $\Theta_0(\delta)$, the denominator is achieved at $\hat\theta_t$, while the numerator is given by the supremum over $\Theta_1(\delta)$. By continuity of the log-likelihood, this supremum is obtained by the same boundary optimization on $d(\theta)=\delta$ approached from the violation side, so the complementary representation is
\begin{equation}
\Lambda_t
=
\sup_{\theta \in \Theta_1(\delta)} \ell_t(\theta)
-
\ell_t(\hat{\theta}_t)
\le 0.
\end{equation}
Thus, the same boundary optimization resolves both acceptance-side and rejection-side computations. The same logic applies to the Gaussian proxy audits after replacing the Bernoulli log-likelihood with the corresponding Gaussian log-likelihood.

These thresholds should therefore be interpreted as operational decision rules rather than exact finite-sample guarantees.

\end{document}